\documentclass[letterpaper, 10 pt, journal]{IEEEtran}  

\usepackage{color}
\newcommand{\ie}{i.e.,\ }
\newcommand{\eg}{e.g.,\ }
\newcommand{\Reffig}[1]{Figure~\ref{#1}}
\newcommand{\Refsec}[1]{Section~\ref{#1}}

\newcommand{\Refeq}[1]{Equation~\ref{#1}}
\newcommand{\Reftab}[1]{Table~\ref{#1}}

\usepackage{graphics} 
\usepackage{epsfig} 
\usepackage{algorithm}
\usepackage{algpseudocode}
\usepackage{amsmath}
\usepackage{amssymb}
\usepackage{subfigure}
\usepackage{graphicx}
\usepackage{subfigure}
\usepackage{booktabs}
\usepackage{threeparttable}
\usepackage{multirow}
\usepackage{multicol}
\usepackage{makecell}
\usepackage{dcolumn}
\usepackage[T1]{fontenc}

\newcolumntype{d}[1]{D{.}{.}{#1}}
                                  
\IEEEoverridecommandlockouts

\title{\LARGE \bf
DL-SLOT: Dynamic LiDAR SLAM and object tracking based on collaborative graph optimization 
}

\author{Xuebo Tian$^{\dagger,1,2}$, Zhongyang Zhu$^{\dagger,1,2}$, Junqiao Zhao$^{*,1,2,3}$, Gengxuan Tian$^{1,2}$, and Chen Ye$^{1,2}$
\thanks{*This work is supported by the National Key Research and Development Program of China (No. 2021YFB2501104)}
\thanks{$^{1}$Department of Computer Science and Technology, School of Electronics and Information Engineering, Tongji University, Shanghai, China}
\thanks{$^{2}$The Key Laboratory of Embedded System and Service Computing, Ministry of Education, Tongji University, Shanghai, China}
\thanks{$^{3}$Institute of Intelligent Vehicles, Tongji University, Shanghai, China}
\thanks{*Corresponding Author:
        {\tt\small zhaojunqiao@tongji.edu.cn}}%
\thanks{$^{\dagger}$Xuebo Tian and Zhongyang Zhu contributed equally to this paper.}%
}

\begin{document}

\maketitle

\begin{abstract}
Ego-pose estimation and dynamic object tracking are two critical problems for autonomous driving systems.
The solutions to these problems are generally based on their respective assumptions, \ie{the static world assumption for simultaneous localization and mapping (SLAM) and the accurate ego-pose assumption for object tracking}.
However, these assumptions are challenging to hold in dynamic road scenarios, where SLAM and object tracking become closely correlated.
Therefore, we propose DL-SLOT, a dynamic LiDAR SLAM and object tracking method, to simultaneously address these two coupled problems.
This method integrates the state estimations of both the autonomous vehicle and the stationary and dynamic objects in the environment into a unified optimization framework. 
First, we used object detection to identify all points belonging to potentially dynamic objects. 
Subsequently, a LiDAR odometry was conducted using the filtered point cloud.
Simultaneously, we proposed a sliding window-based object association method that accurately associates objects according to the historical trajectories of tracked objects. 
The ego-states and those of the stationary and dynamic objects are integrated into the sliding window-based collaborative graph optimization.
The stationary objects are subsequently restored from the potentially dynamic object set.  
Finally, a global pose-graph is implemented to eliminate the accumulated error.
Experiments on KITTI datasets demonstrate that our method achieves better accuracy than SLAM and object tracking baseline methods. 
This confirms that solving SLAM and object tracking simultaneously is mutually advantageous, dramatically improving the robustness and accuracy of SLAM and object tracking in dynamic road scenarios.
\end{abstract}

\section{INTRODUCTION}

\IEEEPARstart{R}{ecently}, LiDAR simultaneous localization and mapping (SLAM) method has been well studied as a fundamental capability in autonomous driving systems. 
Although several advanced LiDAR SLAM methods have been proposed and have high accuracy \cite{loam2014,shan2018lego}, they all build on the static world assumption.

The conventional method is to either filter out point clouds belonging to potential dynamic objects to eliminate the impact of dynamic objects on the SLAM \cite{moving_landmarks,Victor2019,li2019net,demir2019robust} or to separately track and remove objects using traditional multi-object tracking methods \cite{rogersslam,moosmann2013joint,wang2019robust}. 
Although such methods hold the static world assumption, the loss of dynamics information can reduce localization accuracy and cause the SLAM process to fail where the environment is cluttered with moving objects. 

The latest studies coupled SLAM and multi-object tracking for localization in dynamic environments \cite{huang2019clusterslam,yang2019cubeslam,zhang2020vdo,bescos2021dynaslam}.
These methods implement a bundle adjustment (BA) with camera poses, visual features and objects. 
However, 3D object tracking using a camera is imprecise and computationally intensive.

Motivated by these coupled methods, we propose DL-SLOT, which simultaneously realizes LiDAR SLAM and object tracking (SLOT) in dynamic road scenes. 
The state estimations of the ego-vehicle and the stationary and dynamic objects are incorporated into a single graph-based optimization framework.

\begin{figure}
        \centering
        \includegraphics[width=8cm,height=6cm]{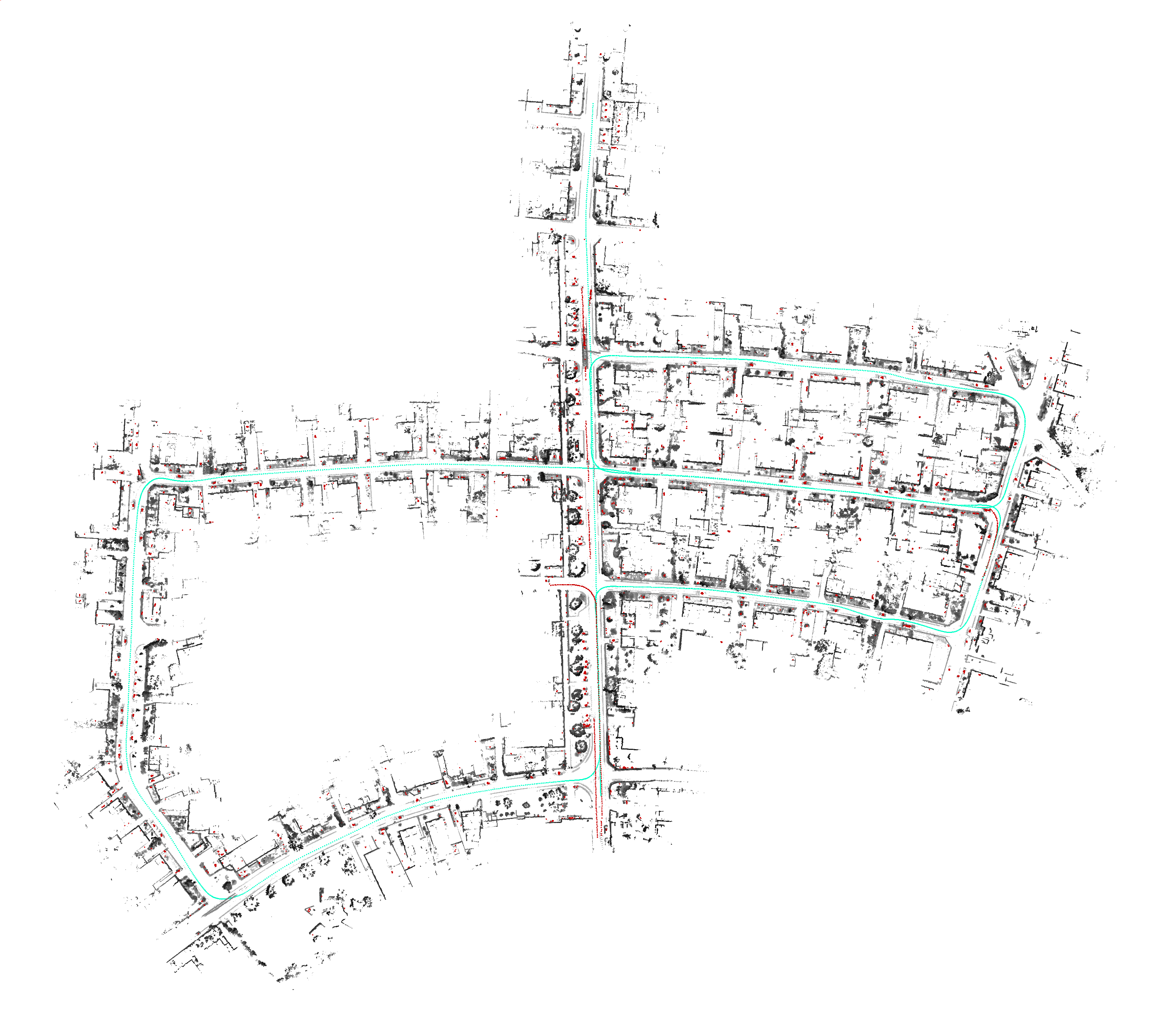}
        \caption{Mapping result of KITTI Odometry dataset 05 sequence. The gray shading indicates the height information of the environment. The green and red dots indicate the ego-trajectory and the trajectories of dynamic objects, respectively.}
\end{figure}

First, an object detector is employed to identify potential dynamic objects, \ie{vehicles, cyclists, and pedestrians}, from each LiDAR frame.
Subsequently, the preliminary LiDAR odometry is implemented using the point cloud that has been filtered out of the points belonging to the potential dynamic objects. 
Meanwhile, we propose using a sliding window with a fixed time interval to perform data association between the current object detection results and the tracked objects using the historical trajectories of the tracked objects, which significantly enhances the association's accuracy.
Finally, assuming that dynamic objects move at a constant velocity for a short period of time, we propose integrating the constant velocity constraint and the states of stationary and dynamic objects and ego-vehicles into a sliding window-based graph optimization framework.

The main contributions of this article are two-fold:
\begin{itemize}
\item A unified optimization framework simultaneously estimating the states of the ego-vehicle and potential dynamic objects.
\item A robust and efficient data association method using trajectory information in a sliding window with a fixed time interval.
\end{itemize}


\section{RELATED WORKS}

\subsection{LiDAR SLAM in dynamic scenes}

Several LiDAR SLAM methods that operate stably in stationary scenes fail in highly dynamic settings, including busy urban roads, because the static assumptions of landmarks do not hold.

\cite{Victor2019} developed a two-stream LiDAR deep convolution network (DCN) that uses the 3D LiDAR point cloud's front and bird's eye view projections as input, to segregate the movable objects from the scene.
After removing the points associated with movable items, the point cloud was used to implement odometry and create a permanent 3D map for localization.
However, points belonging to the stationary objects are also removed, reducing the accuracy of the SLAM.

\cite{chen2019iros} used DCN to obtain the semantic label of each point. 
Based on the semantics, this method constructs a global consistent semantic map that filters out all the points belonging to potential dynamic objects.
Similarly, stationary objects do not contribute to the SLAM optimization.

\cite{DLOAM} projected LiDAR frames into the range image and fed various range frames into DCN for dynamic object detection. 
The method extracts feature points from stationary objects and scenes to compute adjacent frames pose transformation.

In \cite{li2019net}, a dynamic area mask prediction network was proposed to help odometry regression network focus on stationary areas, improving the robustness. 
However, unsupervised DCN trained on scenario-specific datasets has limited generalizability.

In summary, a common feature of the existing dynamic SLAM methods is that the dynamic objects are entirely filtered out; therefore, their contribution to the SLAM system is neglected.

\subsection{SLAM and Object Tracking}

Existing SLOT research is primarily proposed in visual SLAM systems.

\cite{Vincent2020object_tracking} performed instance semantic segmentation of the image using \cite{bolya2019yolact} to obtain the bounding box, class, and binary mask for each instance.
After computing the 3D centroid for each masked object, the states of the objects were estimated by extended Kalman filters and categorized as moving or idle. 
Subsequently, the visual odometry algorithma used a depth image masking the dynamc objects. 
However, the odometry calculation for object tracking ignores all the potential dynamic instances in the imprecise image.

\cite{2020dot} also segmented potential dynamic objects by DCN. 
Subsequently, after estimating the camera motion using stationary region points, the object motion was estimated by minimizing the photometric reprojection error.
Additionally, the dynamic masks were generated for the SLAM system instead of continuously tracking the state of dynamic objects.
However, the camera motion and object motion are not jointly optimized in this system.

\cite{yang2019cubeslam} firstly proposed using bundle adjustment (BA) to jointly optimize ego-poses, states of objects, and feature points. 
Object's cuboid box was generated from 2D bounding boxes and vanishing points sampling. 
Subsequently, these objects were associated by feature point matching, and objects with few associated feature points were considered dynamic.
Rather than considering dynamic objects as outliers, this method optimized the ego-poses and states of objects based on dynamic features and motion model constraints, together with stationary feature observations.
However, the proposed 3D pose recovery method for objects is not robust.

\cite{zhang2020vdo} leveraged optical flow to track the feature points on the objects to achieve object association. 
The motion of the feature points associated with the object was then represented using a motion model. Motion constraints were added to the factor graph to enable the combined optimization of ego-pose and object state. 
In contrast with \cite{yang2019cubeslam}, \cite{zhang2020vdo} performs little marginalization within the estimation window. Therefore, the factor graph constructed by this method is bulky.

According to object uncertainty, observation quality, and prior information, \cite{switch_dynamic_slam} categorized objects into ``good'' and ``bad'' ones.
The measurement constraints of ``good'' dynamic objects and stationary landmarks were leveraged to perform SLAM and object tracking. Those of ``bad'' objects were used for object tracking based on the optimized camera state.
However, object classification needs param tuning, and the additional tracking process is expensive.

\cite{bescos2021dynaslam} made use of instance semantic segmentation and ORB features to track dynamic objects.
Under the constant speed motion model, the objects were associated based on the matched dynamic feature points, and stationary matched features were used to initialize the camera pose. 
Subsequently, a BA solution tightly optimized the scene structure, the camera poses, and the trajectories of objects in a local-temporal window. 
However, the feature-based method reduces its ability to detect accurate bounding boxes and track low-texture objects.

In summary, existing visual SLAM and object tracking methods either suffer from computational inefficiency or are subject to feature degraded scenarios, \eg{texture-less, low illumination}.
Inspired by the collaborative back-end in visual SLAM, this study proposes a precise 3D object tracking and collaborative optimization method for LiDAR SLAM.

\section{METHODS}

\subsection{Problem Formulation}

\subsubsection{Notation definition}
\begin{figure}[ht]
        \centering
        \includegraphics[width=8cm]{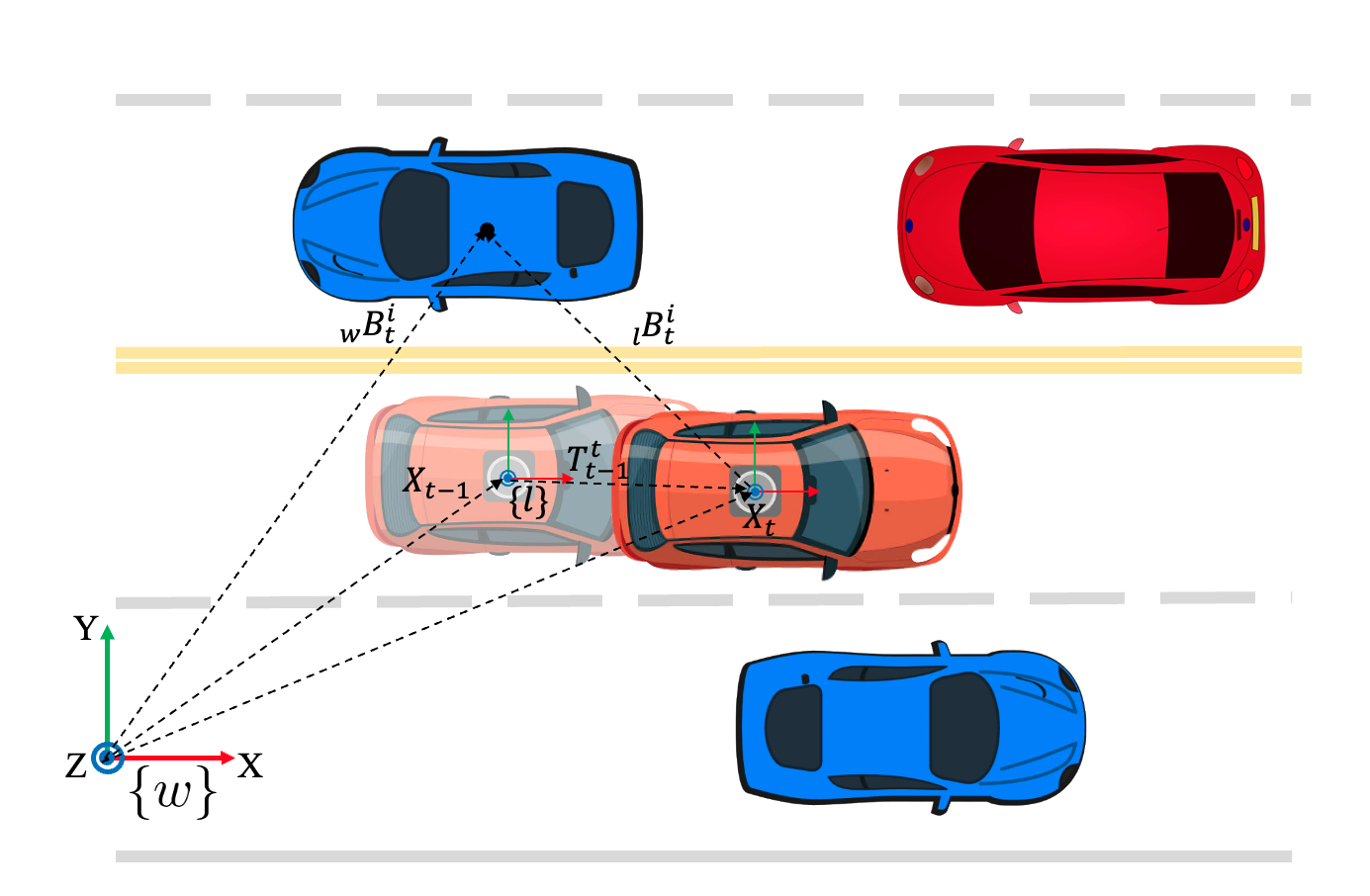}
        \caption{Pose notations of autonomous vehicle and objects. $\{w\}$ is the world coordinate system and $\{l\}$ is the self-vehicle coordinate system. $X$ is the autonomous vehicle's pose, and $T_{t-1}^{t}$ is the odometry between the adjacent frames. $_{w}B_{t}^{i}$ and $_{l}B_{t}^{i}$ are the object's poses in different coordinate systems.} 
        \label{fig_coordinate}
\end{figure}

Two coordinate systems are included in this research: the self-vehicle coordinate system $\{l\}$ and the world coordinate system $\{w\}$, as shown in \Reffig{fig_coordinate}. 
In the world coordinate system, the pose of the autonomous vehicle at the frame $t$ is denoted as $X_{t} \in SE(3)$, and its pose transformation from the previous frame $t-1$ to the current frame $t$ is denoted as $T_{t-1}^{t} \in SE(3)$. 
The detected position and direction of the $i$-th object at time $t$ are $_{l}o_{t}^{i}=(x,y,z)$ and $_{l}yaw \in [-\pi,\pi]$ in $\{l\}$ ,and its positon in $\{w\}$ is $_{w}o_{t}^{i}$, that can be estimated by \Refeq{eq_wo}.
We represent the object's pose as $_{l}B_{t}^{i}$ and $_{w}B_{t}^{i}$ in the coordinate system $\{l\}$ and $\{w\}$, respectively.
\begin{equation}
        {_{w}o_{t}^{i}} =  X_{t} \ast {_{l}o_{t}^{i}}
        \label{eq_wo}
\end{equation}

The first frame is set as the origin of the world coordinate system. The pose transformation sequences of the autonomous vehicle $(T_{0}^{1}, T_{1}^{2}, \cdots, T_{t-1}^{t})$ can be obtained by the existing LiDAR odometry methods. 

The set of objects's position in $\{l\}$ and $\{w\}$ at time $t$ are marked as $_{l}O_{t}=\{_{l}o_{t}^{i}| i \in [1,N] \}$ and $_{w}O_{t}=\{_{w}o_{t}^{i}| i \in [1,N] \}$, where $N$ is the number of objects in $t$-th frame.
The existing tracked object set is $Tr_{t-1} = \{Tr_{t}^{j}| j \in [1,M] \}$, where $M$ is the number of the tracked objects in $Tr_{t-1}$.
The trajectory of $j$-th tracked object at time $t$ is marked as $Tr_{t}^{j}=(_{w}o_{t-n+1}^{j},_{w}o_{t-n+2}^{j}, \cdots ,_{w}o_{t}^{j})$, where $n$ is the length of the trajectory.
The velocity of the $i$-th object from frame $t-1$ to $t$ is involved in the pose transformation matrix $^{i}c_{t-1}^{t} \in SE(3)$.

\subsubsection{Problem definition}
With notations made, the slot problem is defined as the following: given the pose transformation sequence ${T_{0}^{1}, T_{1}^{2}, \cdots, T_{t-1}^{t}}$ and a sequence of object sets ${_{l}O_{0}, _{l}O_{1}, \cdots, _{l}O_{t}}$, simultaneously estimate the autonomous vehicle's poses and the state of each object. 

Specifically, let $\pmb{X}_{L}=\{X_{i}| i \in [0,t] \}$, $\pmb{X}_{O}=\{_{w}B_{j}^{i}| j\in _{l}O_{i} ,i \in [0,t] \}$ and $\pmb{X}_{C}=\{^{j}c_{i-1}^{i}| j\in _{l}O_{i} ,i \in [1,t] \}$.
Given $\pmb{X}=\pmb{X}_{L} \cup \pmb{X}_{O} \cup \pmb{X}_{C}$.
The error terms $e_{odo}$ and $e_{obj}$ are constructed for the ego-vehicle and object poses, and the remaining error term $e_{other}$ is constructed using constraints such as the motion model. 
This problem can subsequently be converted to a least squares problem as follows:
\begin{equation}
        \begin{aligned}
                \pmb{X}^*=\underset{\pmb{X}}{argmin} \{ &\sum_{i\in [0,t]}(\left \| e_{odo}^{i} \right \| ^{2}_{\Sigma_{odo}} + \sum_{ j\in _{l}O_{i}}(\left \| e_{obj}^{j} \right \| ^{2}_{\Sigma_{obj}}  \\
        & + \left \| e_{other}^{j} \right \| ^{2}_{\Sigma_{other}}))
                \}
        \end{aligned}
        \label{eq_prob_def}
\end{equation}

\subsection{System Overview}

\begin{figure*}
        \centering
        \includegraphics[width=18cm]{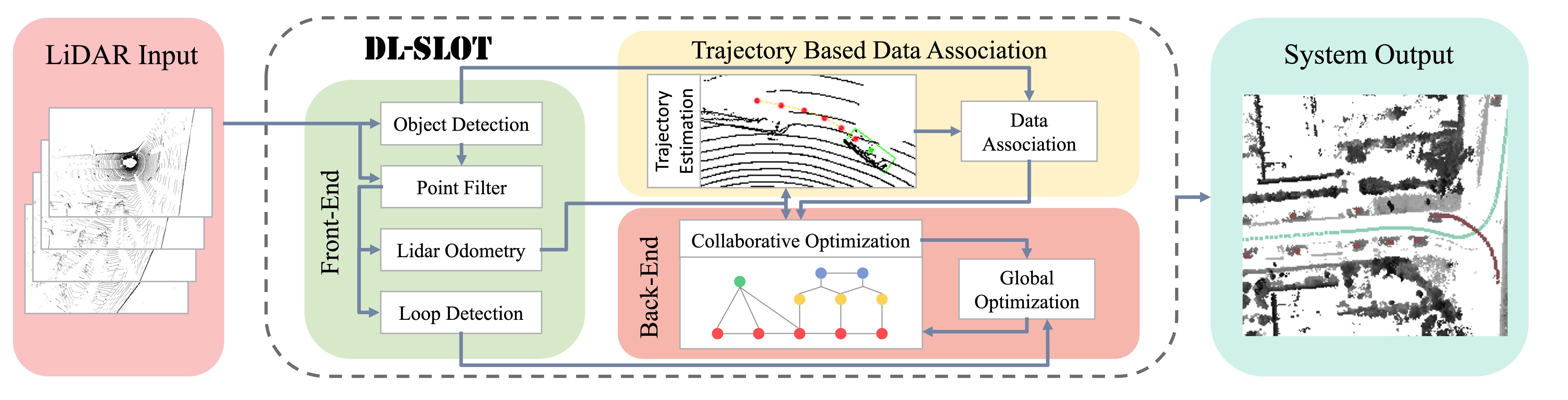}
        \caption{The system architecture of DL-SLOT. The system comprises SLOT front-end, trajectory-based data association, and SLOT back-end. }
        \label{fig_system}
\end{figure*}

The suggested DL-SLOT is a dynamic LiDAR SLAM and object tracking method for outdoor high dynamic scenes, and the system architecture is indicated in \Reffig{fig_system}. 
The system comprises three components: the SLOT front-end, the trajectory-based data association and the SLOT back-end.

The SLOT front-end detects potential dynamic objects using a DCN-based detector \cite{pointrcnn} and filters out LiDAR points in the detection bounding box that may influence the localization accuracy of SLAM.
It subsequently calculates the frame-to-frame transformations according to \cite{shan2018lego}, and detects the loops using \cite{kim2018scan}. 

The data association is performed between the tracked and detected objects using trajectories within a sliding window, which will be elaborated in \Refsec{association_section}. 
The SLOT back-end, comprising sliding window-based collaborative and global optimization, performs the joint optimization of the ego-pose and the states of objects (\Refsec{backend_section}).  

\subsection{Trajectory-based Data Association}
\label{association_section}
Data association has been intensively explored in object tracking \cite{darms2008vehicle,petrovskaya2009model,cesic2014detection}.
\cite{nnda} is a simple and effective data association method, that selects the detected object closest to the predicted position of the tracked object as the associated object. 
\cite{pda} solved the problem of single-target data association in a series of noisy object detections.
\cite{bar1980joint} achieved multi-target data association in cluttered environments, but suffered from a high computational burden.
\cite{mht} dealt with the problem of association conflict in complex motion by forming multiple hypotheses to delay decisions.

\begin{figure}
        \centering
        \includegraphics[width=8cm]{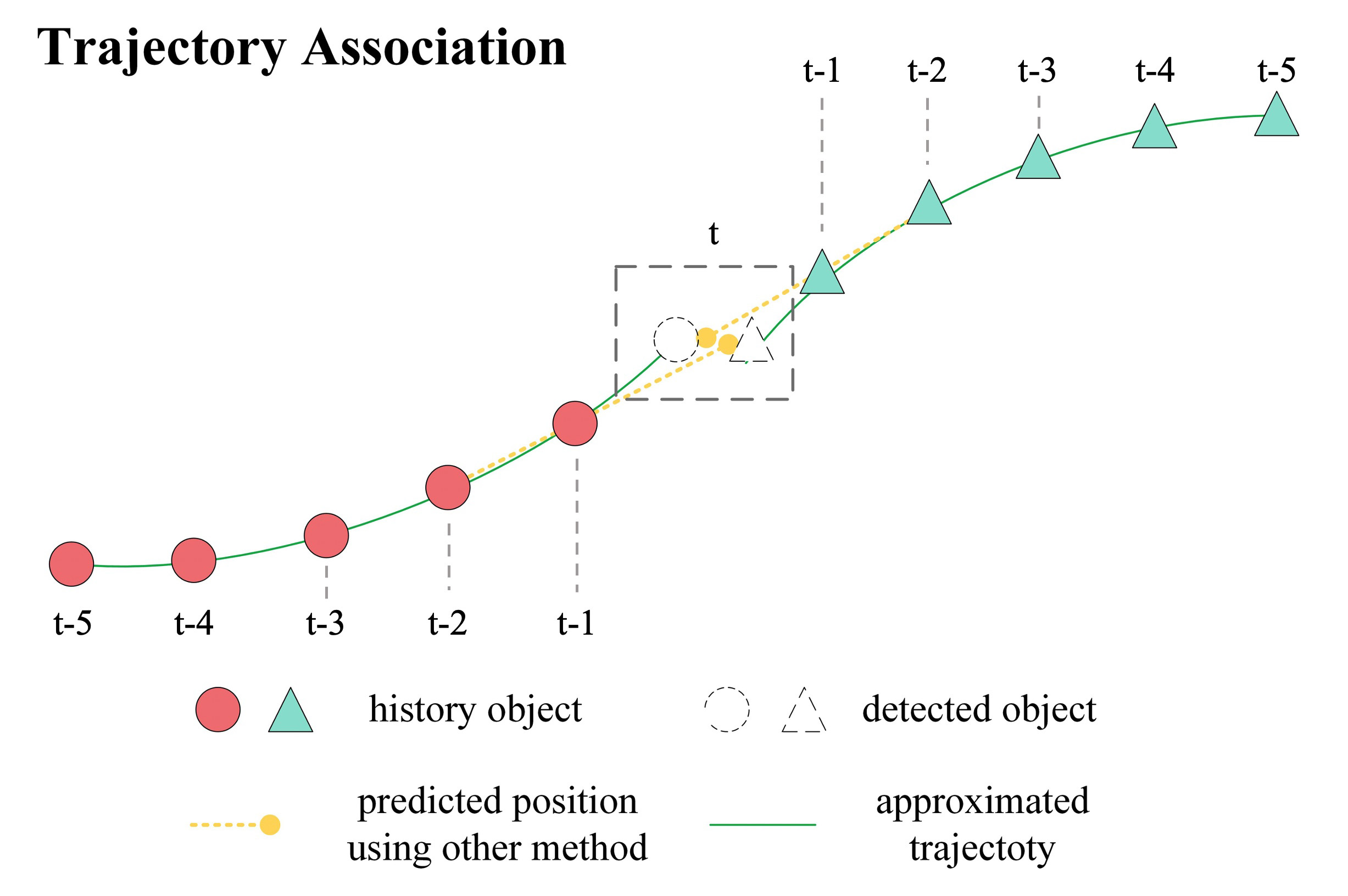}
        \caption{Schematic of trajectory association. The colored circle and triangle denote two different objects. The solid green line represents the approximated object trajectory. The dashed circle and triangle represent the detected objects, and the dashed rectangular indicates a location that is prone to an incorrect association. The yellow dashed line represents the erroneous prediction result.}
        \label{fig_association}
\end{figure}

This study proposes a robust trajectory-based data association, with reliable object detection using LiDAR,
As illustrated in \Reffig{fig_association}, the traditional object tracking methods only use the object information in the adjacent frame, \ie{adopting the assumption of constant velocity motion}; therefore, the predicted results are easily mis-associated, as illustrated by the yellow dashed line. 
The main reason is that the historical states, \ie{trajectories}, of the tracked objects are ignored.
In contrast, the object's current position can be accurately estimated using the approximated object trajectories, as shown by the solid green line in \Reffig{fig_association}.

When an object $_{l}o_{t}^{i}$ is detected at frame $t$, its local pose $_{l}B_{t}^{i}$ is obtained by the object detector, and its global pose $_{w}B_{t}^{i} $ can be calculated by \Refeq{eq_bw}. 

\begin{equation}
        {_{w}B_{t}^{i}} =  X_{t} \ast {_{l}B_{t}^{i}}
        \label{eq_bw}
\end{equation}

Assuming that there are $N$ objects in the set $_{w}O_{t}$ and $M$ trajectories in the existing tracked objects set $Tr_{t-1}$, 
Additionally, the length of the historical trajectory $Tr_{t-1}^{j}$ of each tracked object is $n$, $1 \leq n \leq K-1$, where $K$ is the size of the sliding window which will be discussed later.
We used a polynomial to approximate the historical trajectory $Tr_{t-1}^{j}$ of each tracked object.
The approximation can be achieved by a 3-order polynomial using least-square fitting (\Refeq{eq_apro}) since it can reasonably be assumed that the object moves smoothly over the time interval. 

\begin{equation}
        Pre^{j}(t)=\theta_1 \ast t^{3} + \theta_2 \ast t^{2} + \theta_3 * t + \theta_4
        \label{eq_apro}
\end{equation}
where $\theta_1$, $\theta_2$, $\theta_3$, $\theta_4$ represent the parameters need to be estimated.

We approximated the object's trajectory on the x-axis and y-axis, obtaining the approximate result $Pre^{j}_{x}(t)$ and $Pre^{j}_{y}(t)$. 

The object's position at frame $t$ can be predicted with the approximate result. 
To quantify the degree of matching between the tracked object and the detected object, the matching score matrix $S_t$ is defined, whose dimension is $(M, N)$. 
The element $s_{j, i}$ of $S_t$ indicates the matching score between the $j$-th tracked object and the $i$-th detected object, and the higher score means the higher association possibility between them.
The matching score $s_{j,i}$ was calculated by \Refeq{eq_score}.

\begin{equation}
        s_{j,i}=\left\{
        \begin{array}{ll}
        \multirow{2}*{$  \omega_{j,i} \frac{A-\sqrt{d_{j,i}}}{A}  $}  & (Tr_{t-1}^{j}.init \ \land \ d_{j,i} < \theta_{a}) \\
                                                & \lor \ (\neg Tr_{t-1}^{j}.init \ \land \ d_{j,i} < \theta_{b}) \\
        \specialrule{0em}{0.5ex}{0.5ex}
        0                                       & otherwise \\
        \end{array}\right.
        \label{eq_score}
\end{equation}

\begin{equation}
        \omega_{j,i}=\left\{\begin{matrix}
                1&  &Tr_{t-1}^{j}.cls = {_{w}o_{t}^{i}}.cls \\ 
                0&  &otherwise
               \end{matrix}\right.
        \label{eq_cls}
\end{equation}

\begin{equation}
        d_{j,i}=\left\{
        \begin{array}{ll}
        (Pre^{j}_{x}(t)-{_{w}o_{t}^{i}}.x)^2  &\multirow{2}*{$Tr_{t-1}^{j}.init$}\\
        + (Pre^{j}_{y}(t)-{_{w}o_{t}^{i}}.y)^2&\\
        \specialrule{0em}{0.5ex}{0.5ex}
        ({_{w}o_{t-1}^{j}}.x-{_{w}o_{t}^{i}}.x)^2  &\multirow{2}*{$\neg Tr_{t-1}^{j}.init$}\\
        + ({_{w}o_{t-1}^{j}}.y-{_{w}o_{t}^{i}}.y)^2 &\\
        \end{array}\right.
        \label{eq_dis}
\end{equation}
where $.cls$ denotes the object's class label. 
$\omega_{j,i}$ is the class consistency, and $d_{j,i}$ is the distance between the tracked object and the detected object. 

Since the accuracy of the predicted position of the tracked object with different initialization statuses is different, the initialized status $Tr_{t-1}^{j}.init$ of a tracked object should be assessed (\Refeq{eq_tr_init}).
When an object is observed over a given threshold of frames, its trajectory can be reliably accessed,and marked as initialized.
We set the association ranges $\theta_a$ and $\theta_b$ for the initialized and uninitialized object respectively, and $\theta_a < \theta_b$.
Similarly, only the pose node of the objects that have been initialized will be added to the optimization graph with a sliding window (\Refsec{backend_section}), so that the length of trajectory $n$ in $Tr_{t-1}^{j}$ is smaller than the size of sliding window $K$.

\begin{equation}
        Tr_{t-1}^{j}.init=\left\{\begin{matrix}
                True &  &n > \theta_{init} \\ 
                False&  &otherwise
               \end{matrix}\right.
        \label{eq_tr_init}
\end{equation}

\Refeq{eq_score} indicates the degree of matching between the tracking object and the detected object. 
The closer the predicted and the detected positions are, the higher the matching degree between them is.
After the matching score matrix was constructed, the continuous shortest path algorithm \cite{ahuja1988network} was used to solve the data association problem.

\subsection{DL-SLOT back-end}
\label{backend_section}

\subsubsection{Collaborative optimization}


\begin{figure}
        \centering
        \includegraphics[width=8cm]{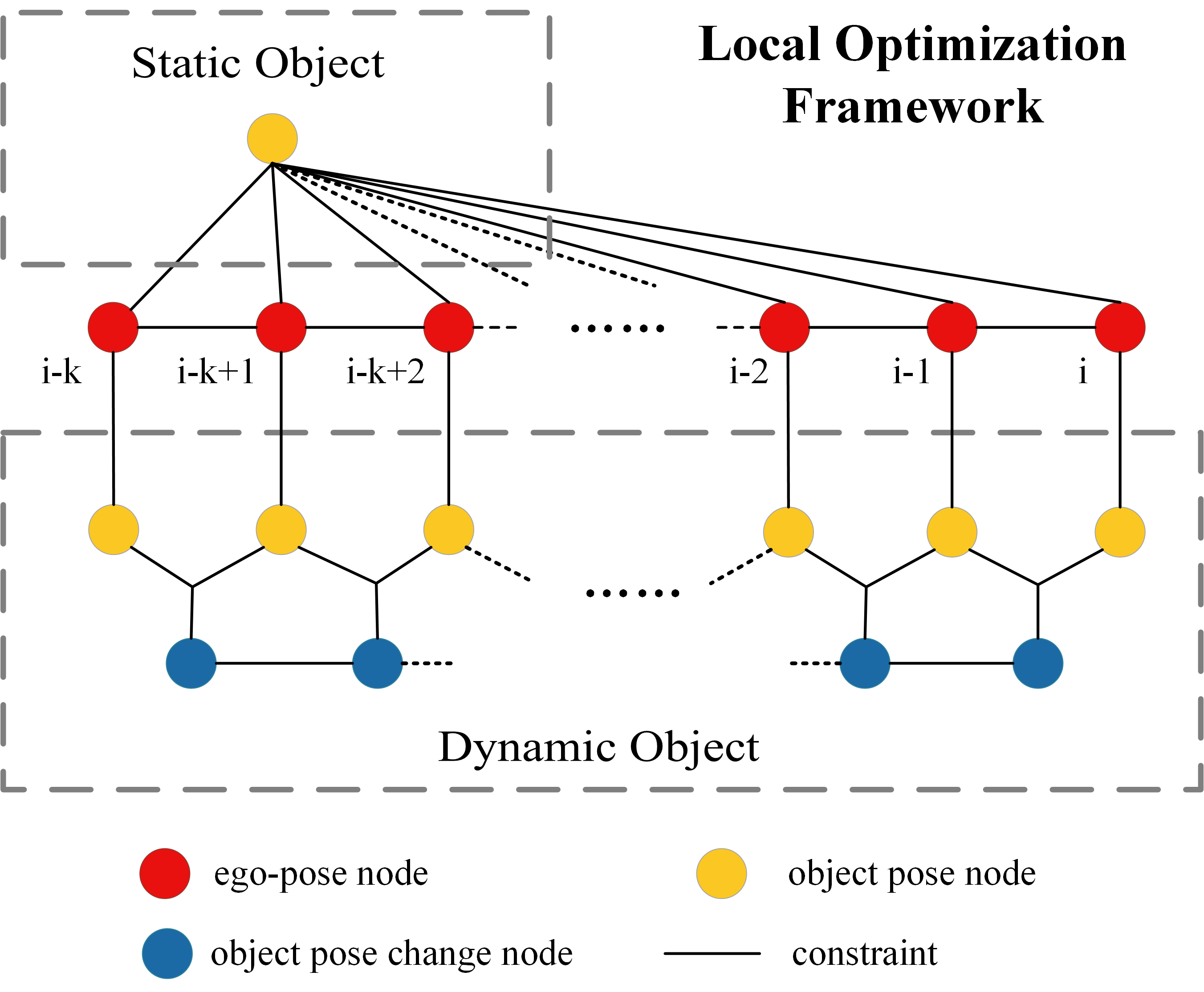}
        \caption{Local optimization framework. The red node represents the ego-pose, and the yellow node represents the detected object pose, and the blue node represents the motion of associated object. The line connecting the nodes is the constructed constraint. }
        \label{fig_local_optimization}
\end{figure}

The back-end is defined based on the graph optimization.
As shown in \Reffig{fig_local_optimization}, all states to be estimated are represented by nodes, and two types of object states: dynamic and stationary, are involved in two dashed boxes. 
The red nodes indicate the Ego-pose $X_{t}$, and the edge between two ego-pose nodes indicates the LiDAR odometry constraint ($e_{odo}$ in \Refeq{eq_e_ego}):
\begin{equation}
        e_{odo}(X_{t-1},X_{t})=({X_{t-1}}^{-1} {X_{t}}) {T_{t-1}^{t}}^{-1}
        \label{eq_e_ego}
\end{equation}

The detected object's global pose is represented by the yellow node in \Reffig{fig_local_optimization}. 
Additionally, the observation constraint $e_{obs}$ between object pose and ego-pose is computed using \Refeq{eq_e_ego_obj}.

\begin{equation}
        e_{obs}(X_{t},_{w}B_{t}^{i}) = ( {{X_{t}}^{-1}} _{w}B_{t}^{i}) { _{l}B_{t}^{i} }^{-1}
        \label{eq_e_ego_obj}
\end{equation}

Furthermore, we added the object motion node $^{i}c_{t-1}^{t}$ into the optimization graph, represented by the blue nodes in \Reffig{fig_local_optimization}, whose initial value $^{i}c_{t-1}^{t}$ is calculated by \Refeq{eq_c}. 
The constraint between object pose and object motion $e_{mov}$ is defined by \Refeq{eq_e_obj_obj}.

\begin{equation}
        ^{i}c_{t-1}^{t} = {_{w}B_{t-1}^{i}}^{-1} \ast _{w}B_{t}^{i}
        \label{eq_c}
\end{equation}

\begin{equation}
        e_{mov}(_{w}B_{t-1}^{i},_{w}B_{t}^{i},^{i}c_{t-1}^{t})= ({_{w}B_{t-1}^{i}}^{-1}  {_{w}B_{t}^{i}}) {^{i}c_{t-1}^{t}}^{-1}
        \label{eq_e_obj_obj}
\end{equation}

We assume a dynamic object moves at a constant speed in a short time period, then the constant speed constraint between the motion node of the associated object $e_{cons}$ is given as \Refeq{eq_e_change_change}. 

\begin{equation}
        e_{cons}(^{i}c_{t-2}^{t-1},^{i}c_{t-1}^{t})= {^{i}c_{t-2}^{t-1}}^{-1} \ast  {^{i}c_{t-1}^{t}}
        \label{eq_e_change_change}
\end{equation}

Finally, our optimization problem is defined by \Refeq{eq_e_all}. 
\begin{equation}
        \begin{aligned}
                \pmb{X}^*=\underset{\pmb{X}}{argmin} \{ &\sum_{i\in [t-K+1,t]} (\left \| e_{odo}(X_{i-1},X_{i}) \right \| ^{2}_{\Sigma_{odo}}  \\
                &+\sum_{j\in ^{init}O_{i} } \left \| e_{obs}(X_{i},_{w}B_{i}^{j}) \right \| ^{2}_{\Sigma_{obs}} \\
                &+\sum_{j\in ^{aso}O_{i}  } \left \| e_{mov}(_{w}B_{i-1}^{j},_{w}B_{i}^{j},^{j}c_{i-1}^{i}) \right \| ^{2}_{\Sigma_{mov}} \\
                &+\sum_{j\in ^{cons}O_{i} } \left \| e_{cons}(^{j}c_{i-2}^{i-1},^{j}c_{i-1}^{i}) \right \| ^{2}_{\Sigma_{cons}} )
                \}
        \end{aligned}
        \label{eq_e_all}
\end{equation}
where $i$ represents the frame's sequence number in the optimization framework, and $\Sigma$ represents the covariance matrix. 
$^{init}O_{i}$ is the set of objects that have been initialized based on \Refeq{eq_tr_init}.
$^{aso}O_{i}$ and $^{cons}O_{i}$ represent the set of objects whose poses have been added to the optimization graph for at least two and three consecutive frames, respectively. 
Therefore, objects in $^{aso}O_{i}$ have at least one motion node, and objects in $^{cons}O_{i}$ correspond to at least one constraint between two motion nodes.
Moreover, $^{cons}O_{i} \subseteq ^{aso}O_{i} \subseteq ^{init}O_{i} \subseteq O_{i}$.

We use a sliding window of size $K$ frames to limit the scale of optimization to reduce the computation and ensure real-time performance. 
When the window slides, the Schur complement is used to preserve the constraints that will be removed from the windows.

\subsubsection{Global optimization}

To further eliminate the accumulated error, we maintained a global pose graph that only contains the LiDAR odometry and loop closure constraints. 
When a loop closure is detected between the $t-d$th frame and $t$th frame, the loop closure constraint is defined by: 

\begin{equation}
        e_{loop}(X_{t-d},X_{t})=({{X_{t-d}}^{-1}} {X_{t}}) {T_{t-d}^{t}}^{-1}
        \label{eq_e_loop}
\end{equation}
where $T_{t-d}^{t}$ is the LiDAR odometry between the $t-d$th frame and $t$th frame.
Therefore, the objective function of the global optimization problem is defined by \Refeq{eq_e_loop_all}. 
\begin{equation}
        \begin{aligned}
        \pmb{X}_{L}^* = \underset{\pmb{X}_{L}}{argmin} \{ &\sum_{i\in [t-d+1,t]}\left \| e_{odo}(X_{i-1},X_{i}) \right \| ^{2}_{\Sigma_{odo}} \\
        &+ \left \| e_{loop}(X_{t-d},X_{t}) \right \| ^{2}_{\Sigma_{odo}}
                \}
        \end{aligned}
        \label{eq_e_loop_all}
\end{equation}

\subsection{Implementation details}

The object detector adopted in this study is PointRCNN \cite{pointrcnn} with the fine-tuned weights on the KITTI dataset.
Three types of potential dynamic objects, \ie{vehicle, pedestrains and cyclists} are detected.
The LiDAR odometry is based on the widely used LeGO-LOAM \cite{shan2018lego}.
And the loop is detected based on Scan Context \cite{kim2018scan}.
Finally, the optimization is implemented based on g2o \cite{grisetti2011g2o}. 



\begin{figure}
        \centering
        \includegraphics[width=8cm]{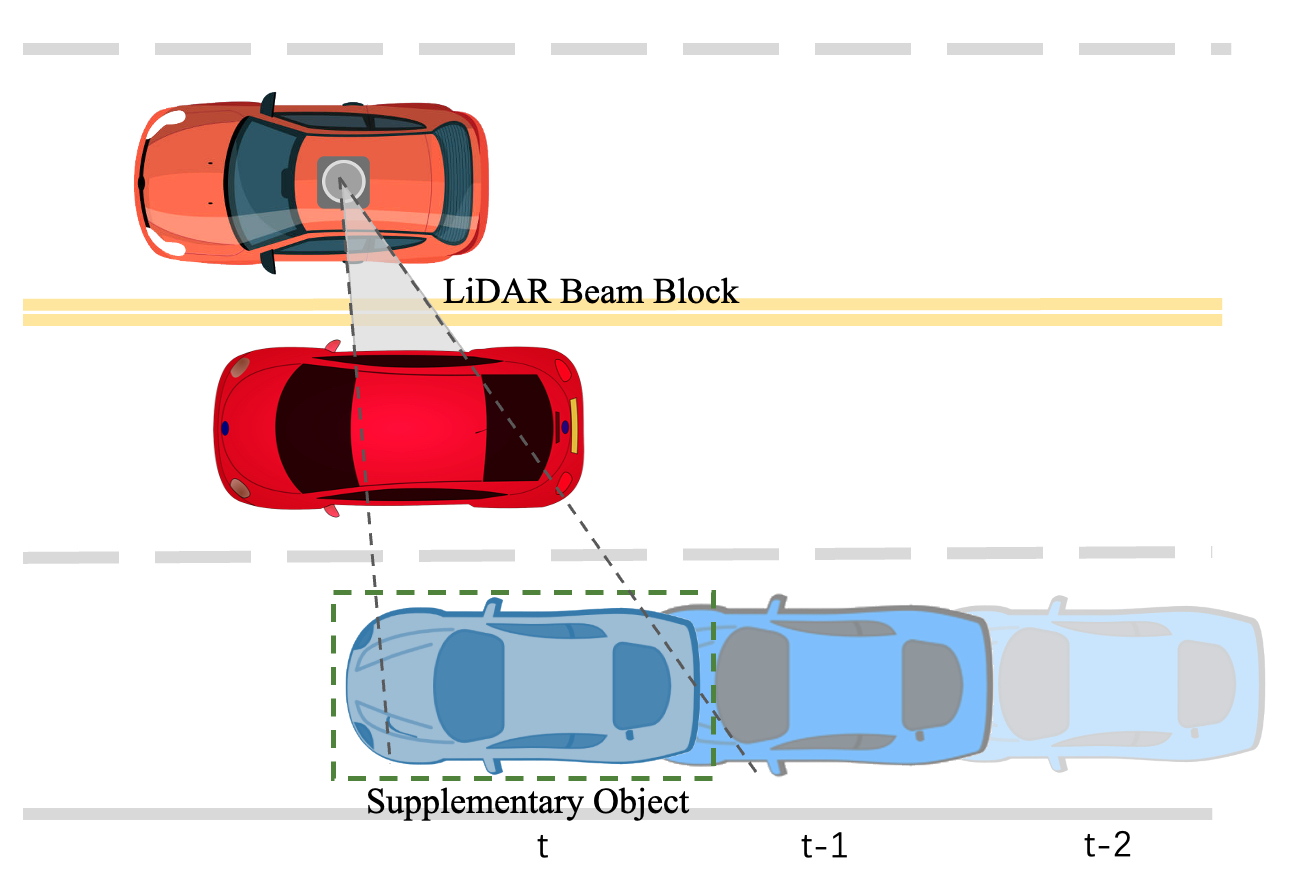}
        \caption{Supplementary detection construction. 
        Since the LiDAR is blocked by the red vehicle, the blue vehicle cannot be detected at time $t$. 
        In this case, the system supplements the missed objects with the predicted position to increase the robustness of the collaborative optimization.}
        \label{fig_occlusion}
\end{figure}

In the trajectory-based data association, if a tracked object is not associated, there are two main reasons: the object is out of the effective sensing range of the LiDAR; the object detection algorithm misses the object due to temporal occlusion. 
To mitigate the influence of the latter situation, we set an upper limit $\theta _{init}^{miss}$ of unassociated frames for initialized tracked objects ($\theta _{init}^{miss}=1$). 
When the number of missed associations is greater than $\theta _{init}^{miss}$, the object is considered out of the sensing range. 
Otherwise, we construct a supplementary detection using the predicted position of the object at frame $t$ and its orientation at frame $t-1$ to facilitate the continued tracking, as shown in \Reffig{fig_occlusion}.

While tracking stationary objects, their global pose might not be stable in continuous observations due to noisy detection or localization.  
Therefore, we added only one object pose node to the collaborative optimization when the object is judged as stationary with a velocity threshold $\theta _{v}=0.1$, which ensures the uniqueness of the stationary object's global position.

However, there may be objects that are not constantly stationary or dynamic when tracked. 
When a static object is judged to be dynamic in this situation, our system will add pose and motion nodes to it and establish constraints.
Conversely, when an object changes from dynamic to static, our system will stop adding new pose nodes to it, thus maintaining its unique global  position.

\section{EXPERIMENTAL RESULTS}



We conducted experiments to evaluate the performance of the proposed method DL-SLOT using the public dataset KITTI \cite{Geiger2012CVPR}. 
The sliding window size $K$ is set as 10, and the threshold $\theta _{init}$ for the tracking initialization is 5.
Additionally, the threshold $\theta_{a}$ and $\theta _{b}$ for data association are set at 1.5 and 3 m, respectively.
Our method is deployed on a PC with ubuntu 20.04, equipped with an Intel Core i7-9800 3.8GHz processor and 32G RAM. 

\subsection{Collaborative Optimization Evaluation}
\label{sec_optimization_eva}

We selected the KITTI Tracking dataset to evaluate the effectiveness of collaborative optimization. 
This dataset was collected in urban areas and highways and contains raw point cloud and GPS data. 
We choose sequences 04, 07, 08, 09, 11, 15, 18, and 19 as the valuation set because these sequences are relatively long.
Furthermore, the driving trajectories contain curved roads rather than a purely straight path.
We adopted the RMSE (Root-Mean-Squared Error) of Absolute Trajectory Error (ATE) to assess the accuracy of trajectories.

For comparison, the widely used LiDAR odometry method, LeGO-LOAM \cite{shan2018lego}, is chosen as the baseline method.
Since LeGO-LOAM did not consider the influence of dynamic objects, we further filtered out the points belonging to potential dynamic objects to generate the LeGO-LOAM* results. 
When only collaborative optimization is performed our method is referred to as DL-SLOT w/o loop; when global optimization is performed, our method is referred to as DL-SLOT. 
Since none of the sequences in the KITTI Tracking dataset contain loop closure, only the results of DL-SLOT w/o loop are given here.


The results are presented in \Reftab{tab_ate_and_rpe}.
In half of the sequences, the localization accuracy of LeGO-LOAM* is lower than that of LeGO-LOAM, because although filtering out the potential dynamic objects avoids the selection of unreliable feature points on moving objects, static objects are also removed, resulting in sparse features and reduced localization accuracy. 
The localization accuracy of our method is higher than that of LeGO-LOAM and LeGO-LOAM* on most sequences, which indicates that the collaborative optimization can effectively improve the localization accuracy, i.e., the autonomous vehicle can optimize ego-pose through continuous observation of the objects' motion.

\begin{table}[ht]
        \caption{ RMSE(m)/RMSE(rad) result of LeGO-LOAM, LeGO-LOAM* and our method on KITTI Tracking dataset.}
        \centering
        \begin{tabular}{ccccccc r@{/}}
        \toprule 
        Seq &  LeGO-LOAM & LeGO-LOAM* & DL-SLOT w/o loop  \\ 
        \midrule 
        04&	1.195/0.269	&1.230/0.264	&\textbf{1.021}/\textbf{0.253}\\
        07&	\textbf{0.821}/0.227	&1.192/\textbf{0.225}	&1.050/0.227    \\
        08&	1.149/\textbf{0.305}	&1.140/0.315	&\textbf{1.131}/0.319   \\
        09&	3.483/0.912	&3.568/0.914	&\textbf{3.271}/\textbf{0.910}\\
        11&	0.206/\textbf{0.247}	&0.247/0.289	&\textbf{0.193}/0.264\\
        15&	0.302/1.485	&0.274/0.516        &\textbf{0.246}/\textbf{0.374}\\
        18&	0.457/0.173	&0.403/0.175	&\textbf{0.408}/\textbf{0.173}\\
        19&	1.572/1.181	&1.566/1.306	&\textbf{1.327}/\textbf{0.563}\\ 
        \midrule
        mean&  1.148/0.600  &1.203/0.501       &\textbf{1.081}/\textbf{0.385}\\
        \bottomrule
        \end{tabular}
        \label{tab_ate_and_rpe}
\end{table}



\subsection{Object Tracking Evaluation}


The assessment of the object tracking is also conducted on the KITTI Tracking dataset.
To evaluate the object tracking results of the proposed method, we use the Precision-Recall metrics and the multi-object tracking accuracy (MOTA) metric.  
The MOTA metric measures the overall accuracy of the object tracking method \cite{bernardin2008evaluating}, and the frame per second (FPS) metric is also given. 
\begin{table}
        \caption{Evaluation result of object detection and object tracking on KITTI Tracking Dataset}
        \centering
        \begin{tabular}{ccccccc}
        \toprule 
        \multicolumn{1}{c}{\multirow{2}{*}{Seq}}        & \multicolumn{2}{c}{Detected Object}& & \multicolumn{3}{c}{Tracked Object}  \\ 
        \multicolumn{1}{c}{}                            & Recall    & Precision               & & MOTA   & Recall    & Precision     \\ 
        \midrule
        04          & 0.9031 & 0.9824    & & 0.8432 & 0.9153 & 0.9460  \\ 
        07          & 0.9834 & 0.9280    & & 0.8923 & 0.9853 & 0.9251  \\ 
        08          & 0.9304 & 0.9899    & & 0.8762 & 0.9317 & 0.9671  \\ 
        09          & 0.9393 & 0.9335    & & 0.8448 & 0.9494 & 0.9205  \\ 
        11          & 0.8913 & 0.9881    & & 0.8571 & 0.9027 & 0.9696  \\ 
        15          & 0.9442 & 0.9694    & & 0.9103 & 0.9558 & 0.9671  \\ 
        18          & 0.8953 & 0.9574    & & 0.8448 & 0.9036 & 0.9466  \\ 
        19          & 0.9855 & 0.9262    & & 0.8718 & 0.9879 & 0.9221  \\ 
        \midrule
        mean        & 0.9341 & 0.9594    & & 0.8676 & 0.9415 & 0.9455  \\
        \bottomrule
        \end{tabular}
        \label{tab_tracking_result}
\end{table}

\Reftab{tab_tracking_result} show the result of object detection and tracking. 
For each tracked object, its intersection of Union (IoU) with ground truth should be above a threshold IoU$_{thres}$ to be considered a successful match.
It can be observed that the Recall of object tracking is higher than that of object detection, which demonstrates the efficacy of the suggested object tracking method and the supplementary detection.

\begin{figure}[ht]
        \centering
        \includegraphics[width=8cm]{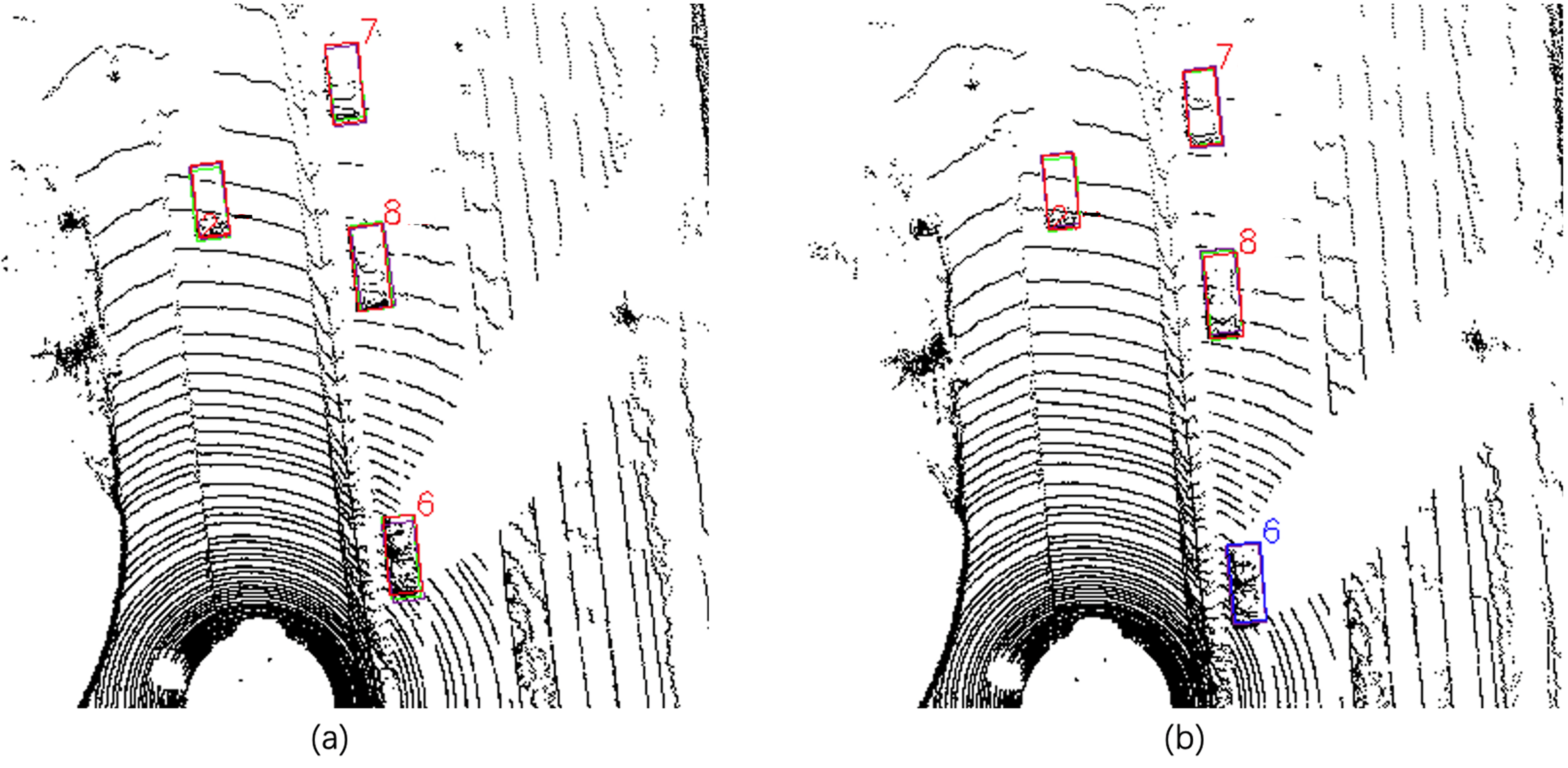}
        \caption{False detection due to lack of the annotation. The red, green and blue bounding boxes indicate the detection, ground truth, and supplemental results, respectively. }
        \label{fig_missing_groundtruth}	
\end{figure}

However, the precision decreases which is primarily due to the omission of the object ground truth annotation in the KITTI Tracking dataset. 
As shown in \Reffig{fig_missing_groundtruth}, the red, green, and blue boxes indicate the detection, the ground truth, and the supplemented objects, respectively. 
The object of ID 6.0 in \Reffig{fig_missing_groundtruth}-(a) is stably tracked, and our method supplements the bounding box of this missing detected object, as shown by the blue bounding box in \Reffig{fig_missing_groundtruth}-(b). 
However, due to the missing annotation in the dataset, this supplemented object is mislabeled as a false detection.

The comparison study is indicated in \Reftab{tab_ab3dmot_compare}.
The baseline method is AB3DMOT\cite{weng2020ab3dmot}, which also uses \cite{pointrcnn} for object detection. 
It combines a 3D Kalman filter and the Hungarian algorithm to implement data association and is the state-of-the-art method in object tracking.
Furthermore, it first proposed to evaluate 3D multi-object tracking with 3D IoU and has high computational effectiveness.

As can be seen from the results, our multi-object tracking performs more than AB3DMOT in terms of comprehensive metric MOTA with different IoU$_{thres}$, even though AB3DMOT uses the groundtruth ego-poses directly. 
Our method has a lower FPS than AB3DMOT, because the object motion estimation and autonomous vehicle pose optimization are conducted concurrently, resulting in higher computing costs. 
However, the processing speed is fully capable of real-time multi-object tracking tasks.

\begin{table*}[ht]
        \caption{Comparison of study of object tracking algorithms on KITTI Tracking dataset.}
        \centering
        \begin{tabular}{ccccc}
        \toprule 
        \multicolumn{1}{c}{\multirow{2}{*}{Method}} &  \multicolumn{3}{c}{MOTA}  & \multicolumn{1}{c}{\multirow{2}{*}{FPS}}  \\ 
        \multicolumn{1}{c}{} & IoU$_{thres}$ = 0.25 &IoU$_{thres}$ = 0.5&IoU$_{thres}$ = 0.7&\multicolumn{1}{c}{} \\
        \midrule 
        AB3DMOT &       0.8624  &       0.8402  & 	0.5706  & 	207.4\\
        ours	&       0.8787  &       0.8676  &	0.6980  &	109 \\
        \bottomrule
        \end{tabular}
        \label{tab_ab3dmot_compare}
\end{table*}

The estimated velocity, pose and trajectory of tracked objects are then examined to further investigate the accuracy of object state estimation through collaborative optimization.

\subsubsection{Evaluation of the Object Pose and Average Velocity}
We selected the dynamic object with the longest tracking length in each sequence and evaluated its trajectory and average velocity. 
The ground truth and the estimated average velocities of the object are marked as v$_{true}$\footnote{v$_{true}$ is obtained by dividing the object ground truth trajectory length during tracking by the tracking time-length.
} and v$_{estimate}$, respectively. 
The results are shown in \Reftab{tab_obj_ate}.
It can be observed that the trajectory accuracy of the tracked objects is great, and the estimated average velocity is likewise near to the ground truth.

\begin{table*}[ht]
        \caption{ Evaluation result of object pose and average velocity on KITTI Tracking dataset.}
        \centering
        \begin{tabular}{ccccccc r@{/}}
        \toprule 
        Seq &  Obj\_id & Tracked frame length & RMSE(m) &RMSE(rad) & v$_{true}$(km/h) & v$_{estimate}$(km/h) \\ 
        \midrule 
        04&	2	&88	&0.119   &0.029  &42.125  &42.264   \\
        07&	6	&61	&0.096   &0.076  &13.608  &13.736    \\
        08&	8	&184	&0.113   &0.023  &46.435  &46.442   \\
        09&	30	&68	&0.150   &0.093  &27.391  &27.749   \\
        11&	0	&372	&0.071   &0.014  &22.661  &22.771   \\
        15&	2	&349    &0.055   &0.186  &9.938  &11.435    \\
        18&	3	&257	&0.093   &0.021  &15.799  &16.088   \\
        19&	63	&131	&0.171   &0.027  &10.613  &10.746   \\ 
        \bottomrule
        \end{tabular}
        \label{tab_obj_ate}
\end{table*}




Additionally, the tracked objects regarded as stationary with tracking lengths longer than 50 frames in the above sequences are shown in \Reftab{tab_static_obj}. 
They have a globally unique pose and zero velocity in our collaborative optimization, providing high-quality observation constraints for ego-pose optimization. 
However, the stationary objects all have a small ground truth average velocity (v$_{true}$) which is due to the noisy annotation. 

\begin{table}[ht]
        \caption{The tracked stationary object information}
        \centering
        \begin{tabular}{cccc}
        \toprule 
        Seq & Obj\_id & v$_{true}$ (km/h) & Tracked frame length  \\ 
        \midrule 
        09	&8     &1.833     &64    \\
        09	&16     &1.909    &29    \\
        11	&28     &1.149    &128   \\
        11	&29      &1.359	  &128    \\
        11	&30	&1.036	 &118    \\
        11	&35     &1.192    &111    \\
        15	&29     &0.648    &349    \\
        19	&33     &1.386    &204     \\        
        \bottomrule
        \end{tabular}
        \label{tab_static_obj}
\end{table}

\subsubsection{Evaluation of the Instantaneous Velocity}

\begin{figure}[ht]
        \centering
        \includegraphics[width=8cm]{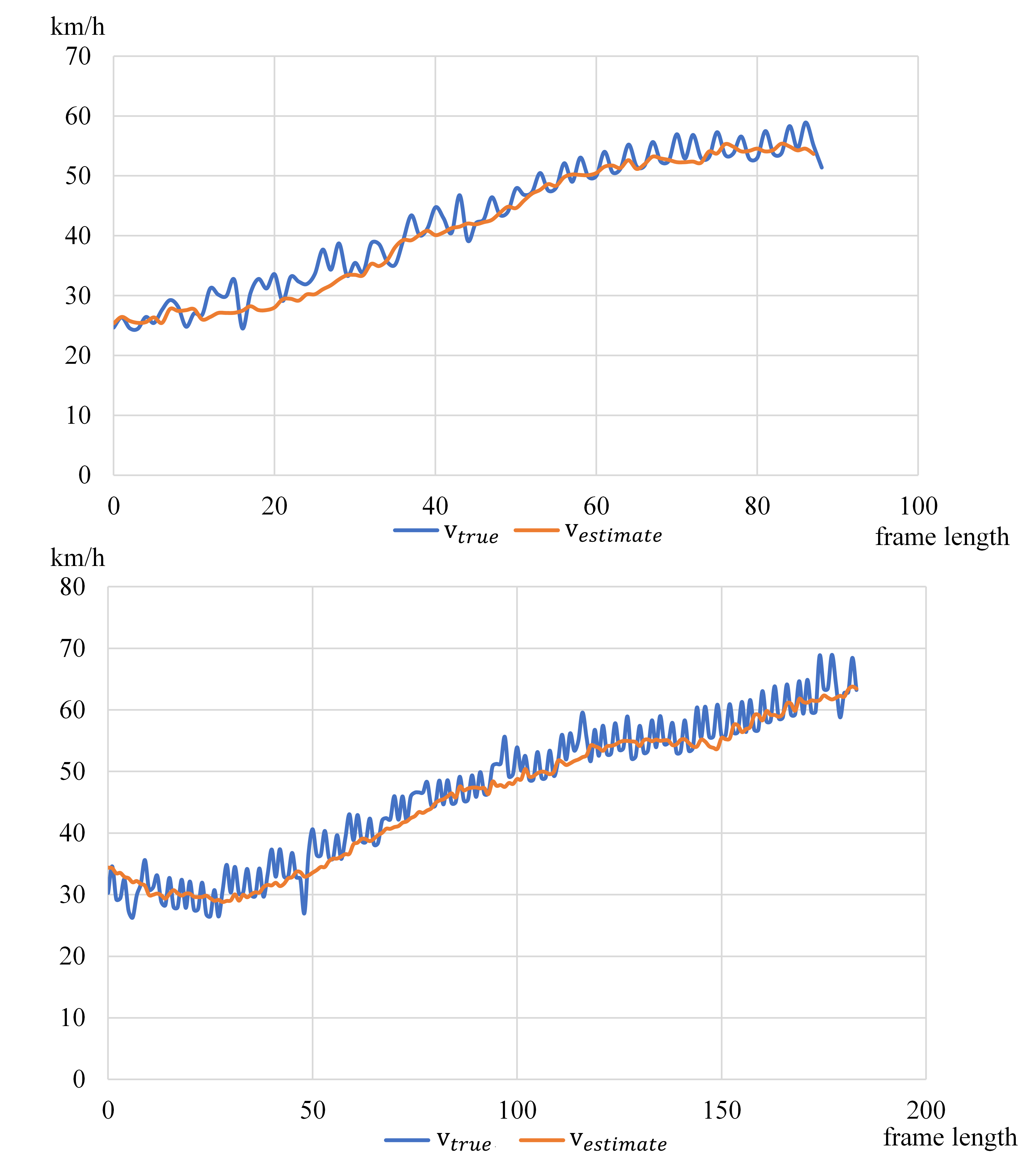}
        \caption{Comparison of the ground truth and the estimated instantaneous velocity. (a) and (b) show the results of object ID 2 in sequence 04 and ID 8 in sequence 08 respectively.}
        \label{fig_ins_v}	
\end{figure}

For dynamic objects, it is crucial to accurately estimate their instantaneous velocities because it affects the object's trajectory prediction. 
We selected two objects with the highest velocity in \Reftab{tab_obj_ate} for evaluation.
As shown in \Reffig{fig_ins_v}, the blue line represents the ground truth instantaneous velocity of an object.
The orange line represents the estimated instantaneous velocity obtained by collaborative optimization.
It can be observed that the ground truth instantaneous velocity fluctuates a lot, which is caused by the low accuracy of object bounding box annotation in the dataset. 
However, the estimated instantaneous velocity is smooth and in the middle of the fluctuation range of the ground truth velocity. 
This result demonstrates that it is feasible to add the constant velocity constraint in collaborative optimization. The object tracking results demonstrate the capability of the proposed method for the short term tracking of dynamic objects. 

\subsubsection{Evaluation of the Pose Prediction}

\begin{figure}
        \centering
        \includegraphics[width=8cm]{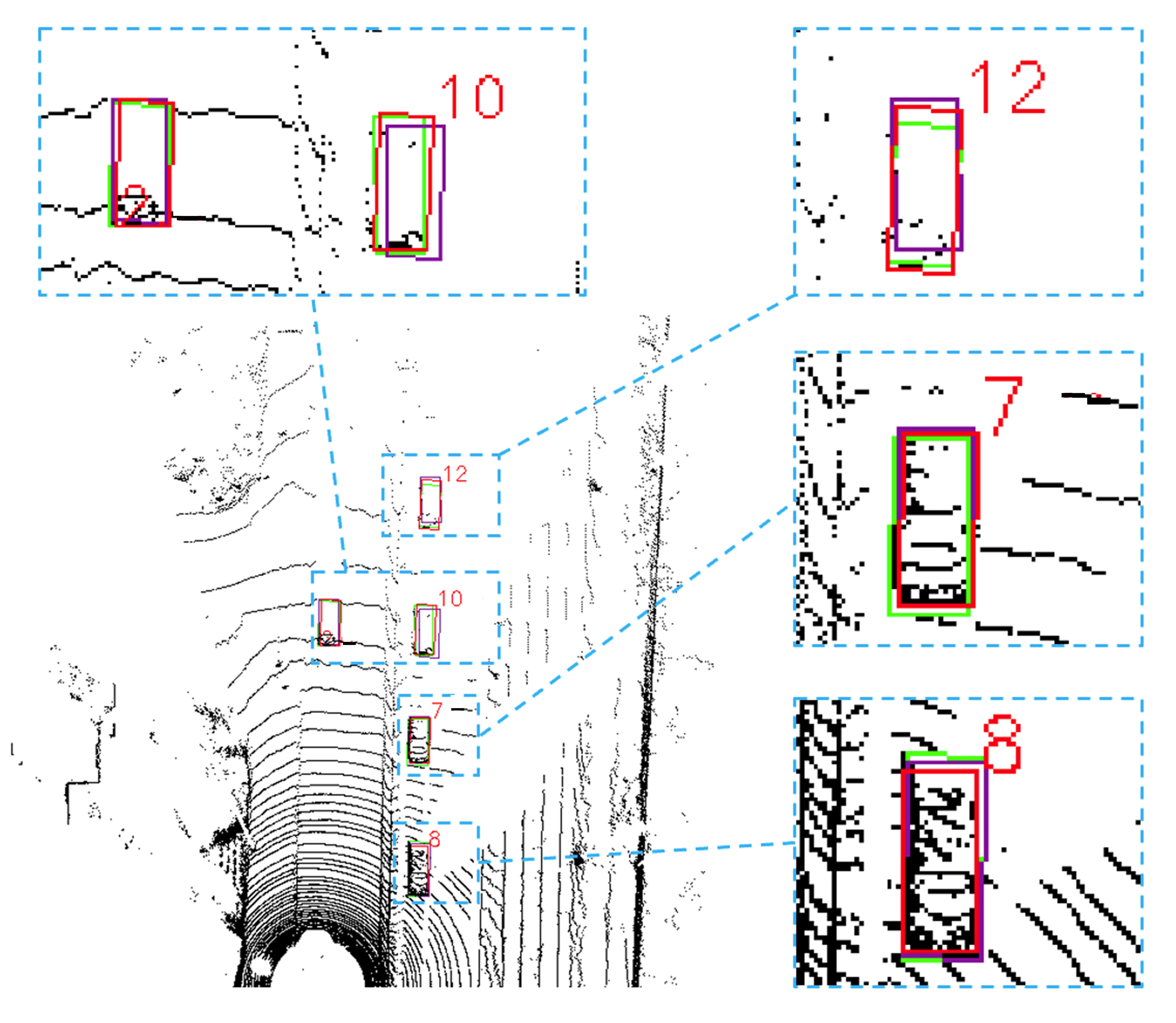}
        \caption{The results of the pose prediction for objects. The green, red and purple bounding boxes represent the ground truth, detected, and predicted pose.}
        \label{fig_predict_position}	
\end{figure}

Accurate pose prediction is crucial for data association. 
\Reffig{fig_predict_position} shows the pose prediction results of dynamic objects in a new frame, in which the green and the red bounding boxes represent ground truth and detected poses respectively. The purple boxes represent the predicted poses of the tracked objects. 
It can be observed that the predicted pose is close to both the ground truth pose and the detected pose. 

\begin{figure}
        \centering
        \includegraphics[width=8cm]{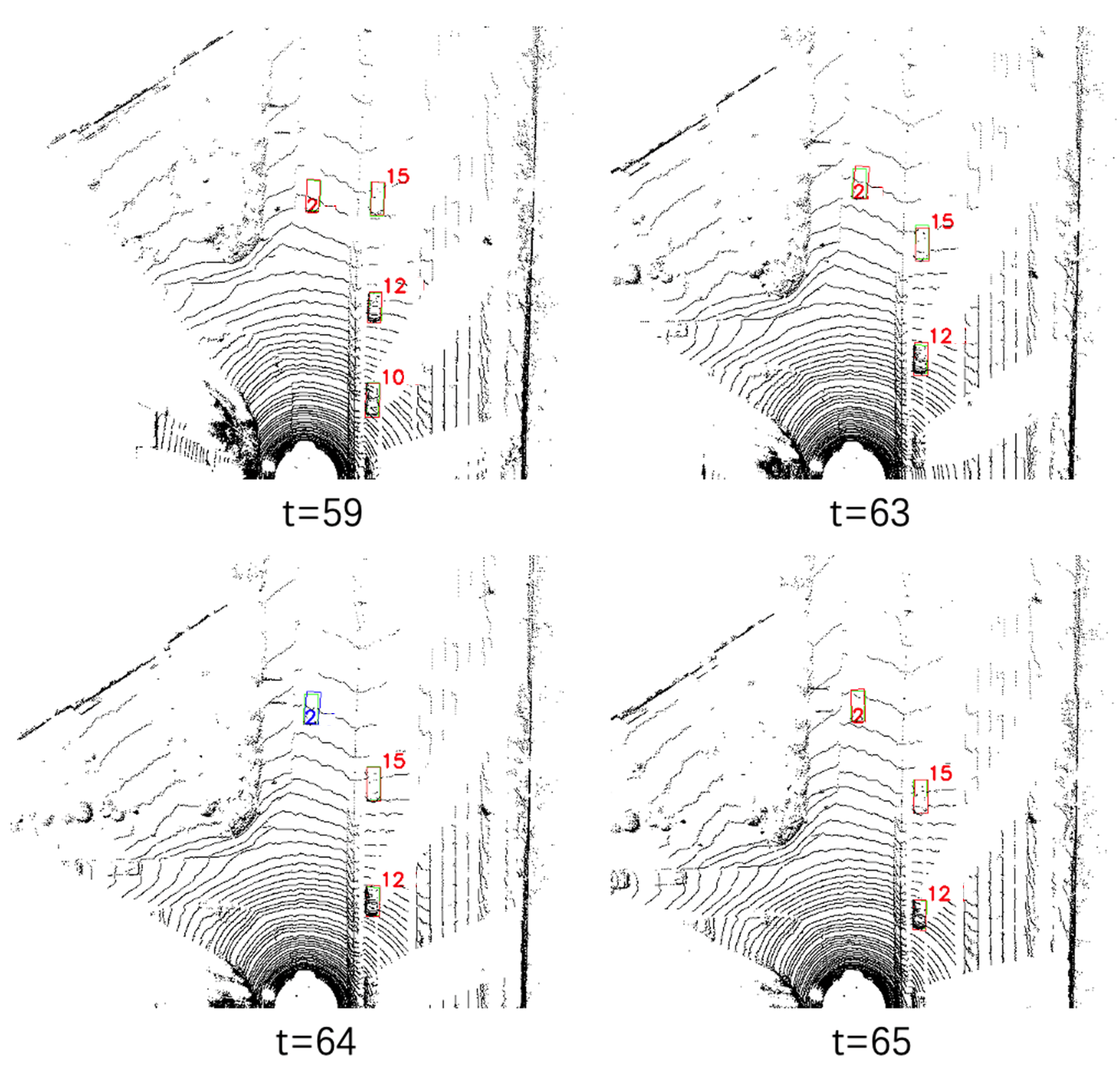}
        \caption{Object tracking result on KITTI Tracking dataset 04 sequence. The green, red, and blue bounding boxes represent the true, detected, and supplementary positions.}
        \label{fig_tracked_result}	
\end{figure}

\Reffig{fig_tracked_result} demonstrates tracking results of sequence 04 from frame 59 to 65. 
Object 2 is detected stably from frame 59 to 63 and at frame 65 as shown by the unique ID, but missed at frame 64.
The suggested object tracking strategy supplements this object at frame 64 based on its predicted pose and completes the association with the trajectory, as shown by the blue bounding box in \Reffig{fig_tracked_result}. 
This shows that the proposed tracking method is robust.

\subsection{Global Optimization Evaluation}


Finally, we conducted experiments on the entire DL-SLOT system that integrates loop closure and global optimization. 
Experiments were performed on KITTI Odometry datasets that contain loops.  
The evaluation metrics are RMSE of ATE, and the compared methods are LeGO-LOAM, LeGO-LOAM*, DL-SLOT w/o loop and DL-SLOT.
The experimental results are shown in \Reftab{tab_system_exp}. 
It can be observed that the localization accuracy of LeGO-LOAM is higher than LeGO-LOAM* in all sequences, because Odometry datasets include more static objects compared to the Tracking datasets, and filtering all movable objects will reduce the localization accuracy.
Additionally, we can find that the DL-SLOT w/o loop has higher localization accuracy than LeGO-LOAM and LeGO-LOAM*, and DL-SLOT has the highest localization accuracy. 
These results suggest that collaborative optimization improves ego-pose accuracy, and global optimization eliminates the accumulated errors generated by odometry and improves the overall accuracy. 

\begin{table*}[ht]
        \caption{RMSE(m)/RMSE(rad) result of LeGO-LOAM, LeGO-LOAM*, DL-SLOT w/o loop and DL-SLOT on KITTI Odometry dataset.}
        \centering
        \begin{tabular}{ccccc}
        \toprule 
        Seq & LeGO-LOAM & LeGO-LOAM* & DL-SLOT w/o loop & DL-SLOT  \\ 
        \midrule 
        00   &13.186/0.457	    &13.539/0.485	&9.746/0.487	&\textbf{2.771}/\textbf{0.029}     \\
        05   &6.753/0.378	&7.682/0.374	&5.604/0.374	    &\textbf{1.834}/\textbf{0.02}    \\
        07   &2.282/0.545       &2.397/0.542	    &2.347/0.547	    &\textbf{1.287/0.017}    \\
        08   &9.300/0.422 	&13.665/0.440	&8.654/0.430	&\textbf{6.576/0.044}   \\
        09   &4.862/0.482	&5.426/0.484	&4.666/0.481	&\textbf{2.764/0.033}    \\   
        \midrule 
        mean &7.277/0.457     &8.542/0.465      &6.203/0.464    &\textbf{3.046/0.029}   \\
        \bottomrule
        \end{tabular}
        \label{tab_system_exp}
\end{table*}

\subsection{Performance Evaluation}

We calculated the average time-consumption of the main functional modules except for object detection, which can be accelerated by GPU. 
As shown in \Reftab{tab_time_consume}, the method proposed in this study enables real-time performance.

\begin{table*}[ht]
        \caption{Average time-consuming of the main functional modules for processing one scan}
        \centering
        \begin{tabular}{cccc}
        \toprule 
        Module & Lidar Odometry & Data Association & Collaborative Optimization  \\ 
        \midrule 
        Average Runtime (ms) &33.6     &9.3    &13.5 \\      
        \bottomrule
        \end{tabular}
        \label{tab_time_consume}
\end{table*}

\section{CONCLUSIONS}

This study proposes an effective and robust LiDAR-based SLOT method that can operate in dynamic scenes. 
This method integrates the state estimation of potential dynamic objects and the autonomous vehicle into a unified collaborative optimization framework. 
Therefore, we can perform SLAM and object tracking simultaneously and make these two processes mutually beneficial. 
Additionally, we introduce an effective trajectory association method, which can accurately predict the tracked object's position using object trajectory.
Our results indicate that DL-SLOT can considerably improve the localization accuracy in dynamic and stationary scenarios and accurately track dynamic and stationary objects, making our method use a wide range of outdoor applications.

In the future, we will investigate a more efficient and robust collaborative optimization framework and tightly couple other measurements from IMU and vision sensors. 

\bibliographystyle{IEEEtran}
\bibliography{reference}

\end{document}